# Deeply Cascaded U-Net for Multi-Task Image Processing

Ilja Gubins[1], Remco C. Veltkamp[1]

[1]Department of Information and Computing Sciences, Utrecht University, Netherlands

{i.gubins[*], r.c.veltkamp}@uu.nl

## Abstract

In current practice, many image processing tasks are done sequentially (e.g. denoising, dehazing, followed by semantic segmentation). In this paper, we propose a novel multi-task neural network architecture designed for combining sequential image processing tasks. We extend U-Net by additional decoding pathways for each individual task, and explore deep cascading of outputs and connectivity from one pathway to another. We demonstrate effectiveness of the proposed approach on denoising and semantic segmentation, as well as on progressive coarse-to-fine semantic segmentation, and achieve better performance than multiple individual or jointly-trained networks, with lower number of trainable parameters.

## 1 Introduction

By now, convolutional neural networks (CNNs) have times and times demonstrated their effectiveness on various tasks. Originally, CNNs were introduced for whole-image classification [Krizhevsky *et al.*, 2012], and later extended for more local tasks such as bounding box object detection [Sermanet *et al.*, 2013] and coarse region-level segmentation [Farabet *et al.*, 2012]. Fully convolutions networks (FCNs) introduced by Long et al. [2015], bridged the gap from coarse to fine, pixel-level, segmentation. One of the key ideas of FCN involved adding a "skip connection" to the architecture: a connection between two or more layers of a neural network that skips one or more layers. Skip connections allowed summation of encoded feature maps from previous layers, therefore enhancing coarse output with fine details.

Ronnenberger et al.'s U-Net [2015] extended FCN by supplementing it's contracting layers with expanding layers where pooling operators are replaced by upsampling operators. In U-Net, skip connections are concatenated and followed by additional convolutions and non-linearities between each depth level, giving the network higher control over feature map combination. This leads to higher resolution, and it was first used for biomedical semantic segmentation, where high accuracy is critical.

[*]Contact author

U-Net architecture and its extensions have proven to be versatile and they have been successfully applied to numerous computer vision tasks: classification, binary and multi-class segmentation, denoising, dehazing and others, in both 2D and 3D (volumetric data and temporal 2D). Practitioners often use complicated multi-stage processing pipelines that include several of these tasks in sequence. However, even if tasks are processed sequentially, it is a common practice to use separate models for each problem, first one neural network for denoising, and then a second for segmentation of the previously denoised data.

An alternative, but conceptually more difficult, approach is to train a multi-task neural network, producing all outputs with one forward pass through the model. This approach involves optimization for multiple tasks at once, and requires the tasks to be related to benefit from parameter sharing. It comes with a number of advantages, the most important of which has been summarized in Caruana [1997], "multi-task learning improves generalization by leveraging the domain-specific information contained in the training signals of related tasks".

In this paper, we propose an approach for multi-task image processing based. More specifically our contributions are as follows:

- We present a novel fully-convolutional neural network architecture U-Net Multi-Task Cascade (UMC), for multi-task learning based on U-Net (Figure 1). We describe architecture building blocks, and propose multiple variations of connectivity: shared encoder connectivity, causal connectivity and densely connected connectivity (Section 3).

- We demonstrate effectiveness of the proposed architecture on two image multi-output processing tasks: joint denoising and semantic segmentation of noisy RGB images (Section 4.1) and coarse-to-fine semantic segmentation (Section 4.2). We compare baseline and multi-stage approaches with our proposed architecture and find that we achieve better performance with lower number of trainable parameters.

## 2 Related work

One of the most notable features of FCN and U-Net architectures is skip connections. In the case of U-Net, skip connec-tions enable the network to propagate spatial information that is lost in pooling operation. Moreover, as found out by Li et al. [2018b], skip (residual) connections make learning easier by smoothing loss landscape. Multiple papers investigated influence of additional skip connections in U-Net architecture, for example short skip connections in encoder blocks [Drozdzal *et al.*, 2016], influence of long skip connections and bridging the semantic gap between connected features [Ibtehaz and Rahman, 2020], as well as additional deeply supervised decoder blocks and creating Nested U-Net [Zhou *et al.*, 2018].

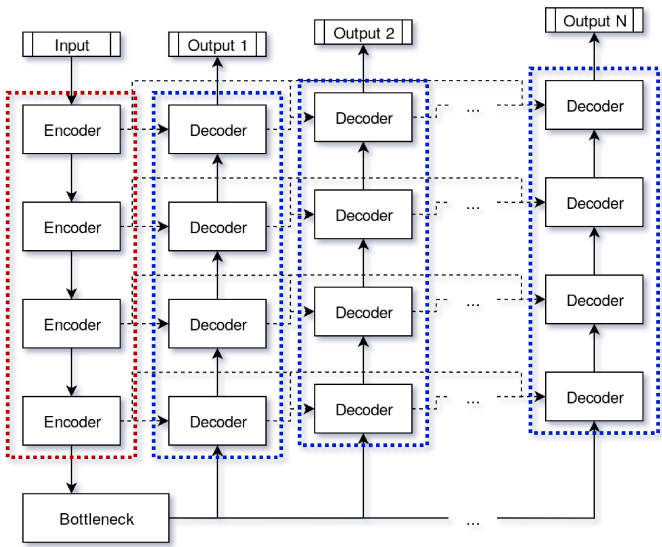
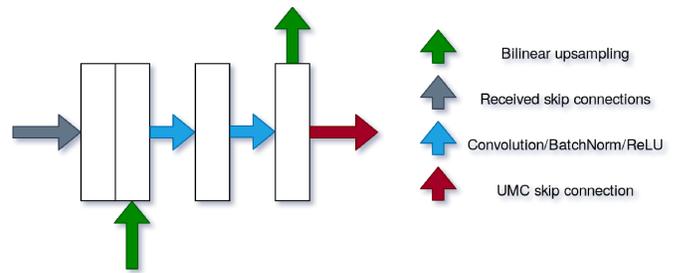

Figure 1: UMC with encoding depth of 4 (highlighted in red) and N decoding pathways (highlighted in blue). Each decoder can receive skip connections from corresponding decoders of previous pathways, while in standard U-Net decoder blocks do not have outgoing skip connections. UMC with one decoding pathway is equal to U-Net.

Figure 2: Single UMC decoder block. UMC skip connection is then concatenated with other skip connections of the same depth level and is propagated forward to the next decoding pathway.

Additional pathways require new connectivity and it is often studied in a practically related task of combining multiple modality inputs. Dolz et al. [2018b] [2018a] proposed using an encoding part of U-Net for each data modality, a single bottleneck and a single decoding U-Net part for intervertebral discs segmentation and ischemic stroke lesion segmentation from multi-modal MRI images. They also explore influence of different connectivity options between encoding pathways and show that related tasks benefit from dense connectivity compared to simpler modality fusion approaches.

As shown by Liu et al. [2018], jointly training cascading networks can improve generalization and improve performance compared to separate training. We can look at this approach as multi-task learning where a single network produces multiple outputs, which brings us to the concepts such as task-related hints [Abu-Mostafa, 1990], deep supervision [Lee *et al.*, 2015] and intermediate concepts [Li *et al.*, 2018a]. All of them describe an idea of adding auxiliary supervision in addition to the overall objective. Such supervision can be used to regularize and guide the network by injecting prior domain structure, leveraging decades of work on topic of organization of computer vision and improving interpretability.

Multiple previously presented works directly extended U-Net for multi-task learning. For example, Sun et al. [2018] proposed concatenating two modified U-Net networks ("stacking blocks") for precise road segmentation on satellite imaging. First block provides auxiliary information, and the second blocks generates road segmentation. Each stacking block is a combination of encoders and decoders similar to U-Net, but with additional outputs from each decoder to facilitate information flow from previous layers. Zhuang et. al. [2018] introduced LadderNet, a similar approach of combining two or more U-Net networks, but instead of concatenating features, LadderNet is summing features. Murugesan et al. [2019] presented Psi-Net, an architecture with a single encoder and three parallel decoders, designed for medical image segmentation. Two of the decoders output auxiliary information which is used to regularize shared encoder with a joint loss, while the third decoder outputs the segmentation map. We advance this idea further to a more general setting of parallel decoders and combine it with insights about connectivity mentioned previously.

## 3 Network architecture

We present U-Net Multi-task Cascade (UMC), a CNN architecture for multi-task learning (Figure 1). We extend U-Net architecture by an additional skip connection in each decoder block (Figure 2). Such outgoing connections allow us add multiple decoding pathways and connect them, forming deep cascades. Accordingly, UMC can be seen as a special case of a multi-task network cascade [Dai *et al.*, 2016] where each cascade stage is a decoding pathway of U-Net.

We hypothesize that connectivity between decoding pathways, and therefore causality, facilitates inductive transfer between early and late stages of cascade. Moreover, such parameter sharing also acts as a form or regularization and reduces the risk of overfitting. We are proposing three options for skip connection connectivity (Figure 3), namely:

1. Shared encoder connectivity, decoders receive only encoder skip connections.

2. Causal cascade connectivity, decoders receive encoder and the previous decoding pathway skip connections.

3. Densely connected cascade, decoders use encoder and

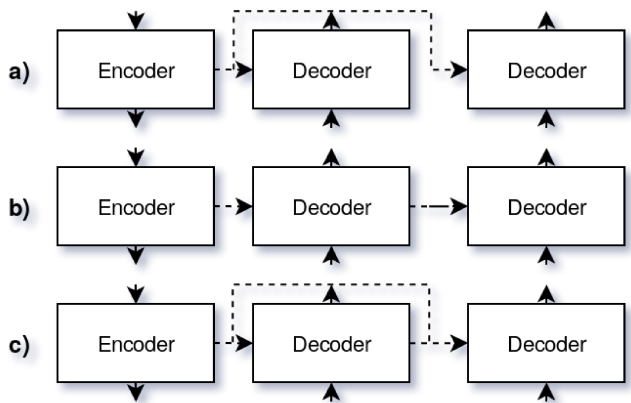

Figure 3: Proposed UMC connectivity configurations: a) shared encoder, b) causal cascade, c) densely connected cascade.

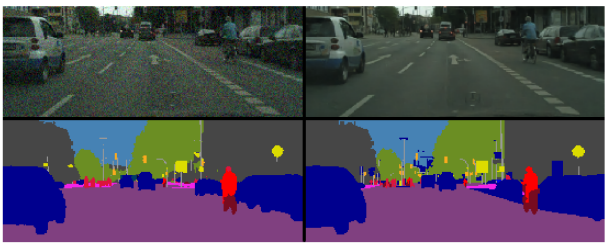

Figure 4: Dense UMC results for noisy images $\sigma = 15$. Top row, left to right: input noisy image, denoised image; bottom row, left to right: segmentation prediction, segmentation ground truth

all (not just the previous) decoding pathway skip connections.

Compared to original U-Net, we have replaced transposed convolution layers with bilinear upsampling to reduce the number of parameters.

# 4 Experiments

We conduct series of experiments to demonstrate effectiveness of UMC in various scenarios and for different data modalities.

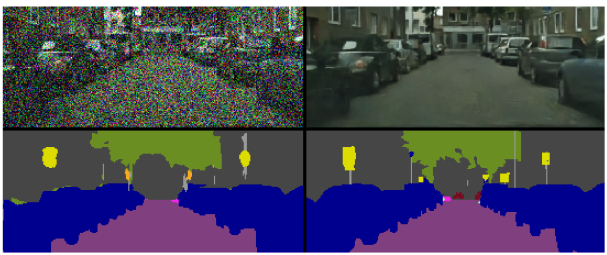

Figure 5: Dense UMC results for noisy images $\sigma = 60$. Top row, left to right: input noisy image, denoised image; bottom row, left to right: segmentation prediction, segmentation ground truth

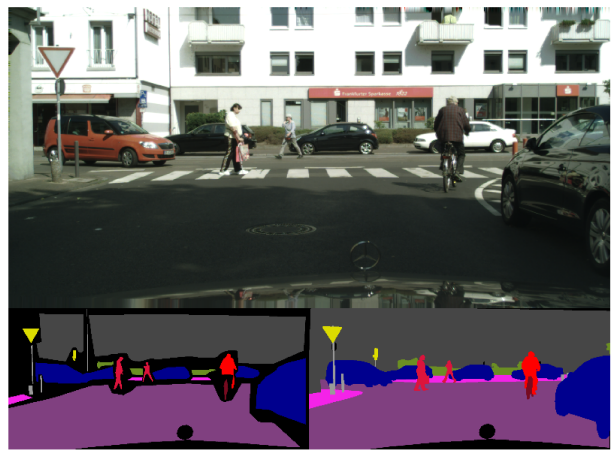

Figure 6: Cityscapes dataset contains high-resolution street imagery (top row) with coarse (bottom left) and fine (bottom right) annotations for 30 classes of 8 categories.

## 4.1 Supervised denoising and semantic segmentation

Semantic segmentation is a task of assigning a label to each pixel. However, when a high-level task such as semantic segmentation is conducted on noisy data, typically an extra preprocessing step of restoration is required to obtain valid results further down the pipeline. Moreover, noise patterns and perturbations can lead to severe misclassifications [Wu *et al.*, 2017]. We hypothesize that our proposed network architecture can incorporate such preprocessing to the network, leading to increased robustness and higher performance.

In this experiment, we compare different configurations of denoising and semantic segmentation networks, at various noise levels. We use Cityscapes [Cordts *et al.*, 2016] dataset, containing 5000 finely annotated images (Figure 6). The images are high-quality urban scenes captured from a road vehicle, annotated into 30 classes, out of which 19 are used for evaluation. During training, we add zero-mean Gaussian noise ($\sigma = 15, 30, 45, 60$) to the Cityscapes images (Figure 4, 5), and then we train and evaluate multiple configurations of U-Net and UMC networks. For evaluation of semantic segmentation performance we use mean intersection over union (mIoU) and for denoising evaluation we calculate peak signal-to-noise (PSNR).

In total, we have evaluated four baseline and multi-stage configurations:

1. Noisy Segmentation U-Net: noisy images are used as the input to a U-Net model that produces semantic segmentation. We use this configuration as a baseline.

2. CBM3D + Clear Segmentation U-Net: we first denoise images with CBM3D algorithm [Dabov *et al.*, 2007], and then feed denoised images to a U-Net model trained to segment noiseless images. This configuration represents one of the most popular and accessible workflows for segmentation of noisy images, as CBM3D is a widely available non-learning based denoising algorithm.

3. Denoising + Segmentation U-Nets: we separately train

a denoising U-Net model and a semantic segmentation U-Net model, and use them sequentially. First noisy images are fed into denoising network, and then denoised images are segmented with a segmentation network.

4. Jointly trained Denoising + Segmentation U-Nets: similar to the previous configuration, we use two U-Net networks, but we employ a joint training approach described by Liu et al. [2018]. The networks are trained simultaneously with a sum (joint loss with $\alpha = 1$) of denoising (reconstruction) and segmentation losses.

To evaluate UMC performance for this application, we train all three types of UMC models with 2 decoding pathways, one for denoising and another for semantic segmentation. To keep the comparison as equal as possible, we use the same depth and number of filters for all U-Net and UMC models. We fix and use the same optimization hyperparameters for all of the models we trained: we used Adam optimizer with learning rate of 0.003, denoising supervision is done with mean squared error loss function and semantic segmentation is facilitated with softmax cross-entropy loss function. We make use of data pre-processing and simple training-time augmentation: the data is normalized over whole dataset mean and standard deviation, cropped to 512x256 pixel resolution, as well as augmented by random horizontal flipping. All of the networks for this experiment were trained for 150 epochs, even if training converges at earlier epochs. The models use same number of filters at each depth level (32, 64, 128, 256, 512).

Results of baseline and multi-stage approaches (Table 1) show that two-network models have better performance than the baseline U-Net and U-Net with CBM3D denoiser on higher noise levels ($\sigma = 45, 60$). Denoising and segmentation combination produces on average higher PSNR for noisier images, while jointly trained denoising and segmentation combination outputs have better segmentation quality.

Results of UMC models (Table 2) show shared encoder connectivity UMC models achieve better denoising quality, while densely connected UMC achieves better quality of segmentation. We hypothesize that shared encoder connectivity limits the semantic gap between encoders and decoders of the same depth, while in densely connected UMC, network finds a way to include the denoised information to improve segmentation.

Comparing results together, UMC models achieve comparable denoising performance and better segmentation performance than separately or jointly trained networks, with lower number of trainable parameters.

### 4.2 Coarse-to-fine semantic segmentation

Training model to produce multiple progressively harder outputs can be viewed as a giving the network task-related hints [Abu-Mostafa, 1990]. This enables guiding of training with more domain knowledge. For example, As Chi Li et al. [Li *et al.*, 2018a] points out, knowing object orientation is a prerequisite to inferring object partial visibility, which in turn constrains the 3D locations of semantic object parts. We hypothesize that our proposed network architecture is well-suited for handling coarse-to-fine progressively harder tasks,

| $\sigma$ | U-Net | CBM3D + CS-U-Net | D+S U-Nets | JT D+S U-Nets |
|---|---|---|---|---|
| 15 | **43.58** (27.78dB) | 10.95 (20.46dB) | 41.87 (**39.48dB**) | 42.11 (38.15dB) |
| 30 | **40.89** (21.76dB) | 7.46 (20.31dB) | 37.91 (**36.75dB**) | 38.62 (34.92dB) |
| 45 | 35.96 (18.24dB) | 5.12 (19.96dB) | **36.28** (**34.71dB**) | 36.25 (33.87dB) |
| 60 | 33.1 (15.75dB) | 3.97 (19.5dB) | 31.64 (**33.52dB**) | **33.32** (32.37dB) |
| P# | 7.766M | 7.766M | 15.532M | 15.532M |

Table 1: Average segmentation (mIoU) and denoising (PSNR) results of baseline and multi-stage approaches on Cityscapes validation dataset. Last row shows rounded total number of trainable parameters for each of the approach. Since U-Net (second column) does not conduct any denoising, denoising results show PSNR of noisy input images. Best results in each row are highlighted in bold.

| $\sigma$ | Shared encoder UMC | Causal UMC | Dense UMC |
|---|---|---|---|
| 15 | 45.31 (**39.19dB**) | 46.30 (38.65dB) | **46.46** (38.21dB) |
| 30 | 41.03 (**36.49dB**) | 41.27 (35.95dB) | **41.33** (35.93dB) |
| 45 | 36.67 (34.19dB) | 37.42 (**34.26dB**) | **38.29** (34.12dB) |
| 60 | 34.85 (**33.29dB**) | 34.49 (32.83dB) | **35.25** (32.69dB) |
| P# | 10.985M | 10.985M | 11.769M |

Table 2: Average segmentation (mIoU) and denoising (PSNR) results of our proposed approaches on Cityscapes validation dataset. Last row shows rounded total number of trainable parameters for each of the approach. Best results in each row are highlighted in bold.

| Approach | $f_{cls}$ acc. (%) | $f_{cls}$ mIoU | P# |
|---|---|---|---|
| Group 1 ($f_{cls}$) | | | |
| UNet | 92.83 | 53.62 | 7.766M |
| Group 2 ($c_{cls}, f_{cls}$) | | | |
| Shared UMC | 93.08 | 53.28 | 10.986M |
| Causal UMC | **93.15** | **53.65** | 10.986M |
| Dense UMC | 93.13 | 53.06 | 11.769M |
| Group 3 ($c_{cat}, f_{cls}$) | | | |
| Shared UMC | 93.19 | 52.53 | 10.985M |
| Causal UMC | 93.09 | **53.89** | 10.985M |
| Dense UMC | **93.21** | 53.72 | 11.769M |
| Group 4 ($c_{cat}, f_{cat}, f_{cls}$) | | | |
| Shared UMC | 92.99 | 51.66 | 14.121M |
| Causal UMC | 92.95 | 50.1 | 14.121M |
| Dense UMC | **93.33** | **52.45** | 16.471M |

Table 3: Average pixel accuracy, mean intersection over union and rounded total number of parameters for each of the described approaches for the second experiment on Cityscapes validation dataset. Best results in each row are highlighted in bold.

as each U-Net Cascade decoding pathway can be used to output and subsequently output one task each.

To evaluate our hypothesis we train UMC models with all previously mentioned types of connectivity. We use same Cityscapes dataset [Cordts et al., 2016] as in previous experiment, but with addition of coarse annotations and hierarchical class categories (Figure 6). For example, category "vehicle" represents multiple classes, such as "vehicle car", "truck", "bus", "bicycle" and others. In total, dataset provides 8 categories for 30 classes. We follow the exact same training protocol as in the previous experiment (Section 4.1).

We group our experiments by the outputs they are supervised to produce from the input image:

1. Fine per-class semantic segmentation ($f_{cls}$)
2. Coarse per-class, fine per-class semantic segmentation ($c_{cls}, f_{cls}$)
3. Coarse per-category, fine per-class semantic segmentation ($c_{cat}, f_{cls}$)
4. Coarse per-category, fine per-category, fine per-class semantic segmentation ($c_{cat}, f_{cat}, f_{cls}$)

The results (Figure 3) suggest that coarse-to-fine spatial tasks benefit from having causality of outputs in the network (group 2). We can also observe that if networks are given tasks that are varying both in spatial resolution (fine/coarse) and categorical resolution (per-category/per-class) information, they benefit from dense connectivity (group 4).

## 5 Conclusion

In this paper we have introduced a convolutional neural network architecture for multi-task learning called U-Net Multi-Task Cascade (UMC). An additional skip connection in decoder blocks allows us to chain decoders into separate output pathways, which we can use for multi-task learning. We have proposed three types of connectivity and evaluated them in two segmentation experiments. First experiment show that for joint denoising and segmentation UMC achieves better performance with lower number of trainable parameters, and consequently faster training. Second experiment provided insight into connectivity in UMC architecture and what kind of tasks might benefit from differences in it.

There are several directions for future research that may improve the UMC performance. First of all, continuing experimentation and exploring connectivity and task causality within a single neural network. Segmentation experiment 4.1 shows a noticeable semantic gap between encoders and decoders of the same depth, and continuing work on understanding should be beneficial for such densely interconnected network as UMC. We also believe that the architecture can be used for more tasks than we have explored in experiments.